\documentclass[runningheads]{llncs}
\usepackage{graphicx}
\usepackage{makecell}
\usepackage{pdflscape,array,booktabs}
\usepackage{enumitem}
\usepackage{outlines}
\usepackage{tabularx}
\usepackage{longtable}
\usepackage{mdframed}
\usepackage{soul}
\usepackage{xurl}
\usepackage{multirow}
\usepackage{subcaption}
\usepackage[textsize=scriptsize,backgroundcolor=yellow!40]{todonotes}

\begin{document}
\title{FLRA: A Reference Architecture for \\Federated Learning Systems}
%
%
\author{Sin Kit Lo\inst{1,2}\orcidID{0000-0002-9156-3225} 
\and
Qinghua Lu\inst{1,2}\orcidID{0000-0002-7783-5183}
\and Hye-Young Paik\inst{2}\orcidID{0000-0003-4425-7388} 
\and Liming Zhu\inst{1,2}\orcidID{0000-0001-5839-3765}}
\authorrunning{SK. Lo et al.}
%
\institute{Data61, CSIRO, Sydney, Australia \and
University of New South Wales, Sydney, Australia}
\maketitle              
\begin{abstract}
Federated learning is an emerging machine learning paradigm that enables multiple devices to train models locally and formulate a global model, without sharing the clients' local data. A federated learning system can be viewed as a large-scale distributed system, involving different components and stakeholders with diverse requirements and constraints. Hence, developing a federated learning system requires both software system design thinking and machine learning knowledge. Although much effort has been put into federated learning from the machine learning perspectives, our previous systematic literature review on the area shows that there is a distinct lack of considerations for software architecture design for federated learning. In this paper, we propose FLRA, a reference architecture for federated learning systems, which provides a template design for federated learning-based solutions. The proposed FLRA reference architecture is based on an extensive review of existing patterns of federated learning systems found in the literature and existing industrial implementation. The FLRA reference architecture consists of a pool of architectural patterns that could address the frequently recurring design problems in federated learning architectures. The FLRA reference architecture can serve as a design guideline to assist architects and developers with practical solutions for their problems, which can be further customised.


\end{abstract}
\section{Introduction}

\begin{figure}
\centering
\includegraphics[width=0.7\textwidth]{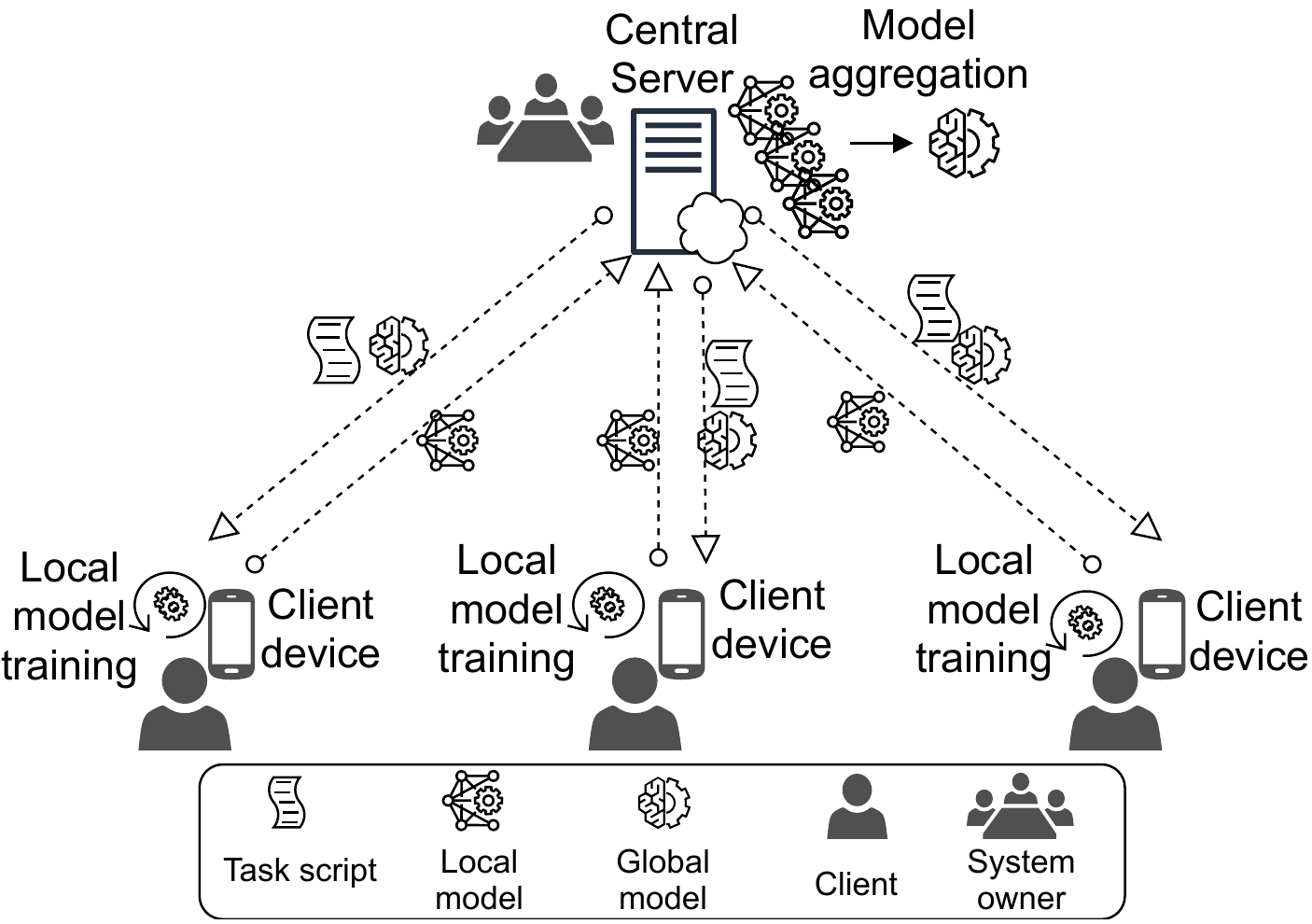}
\caption{Federated Learning Overview~\cite{lo2021architectural}.} \label{FL_Overview}
\end{figure}

The ever-growing use of industrial-scale IoT platforms and smart devices contribute to the exponential growth in data dimensions~\cite{s19204354}, which, in turn, empowers the research and applications in AI and machine learning. However, the development of AI and machine learning also significantly elevates data privacy concerns, and General Data Protection Regulation (GDPR)\footnote{\url{https://gdpr-info.eu/}} stipulates a range of data protection measures with which many of these systems must comply. This is a particular challenge in machine learning systems as the data that is ready for model training is often insufficient and they frequently suffer from ``data hungriness issues''. As data privacy is now one of the most important ethical principles of machine learning systems \cite{jobin2019global}, there needs to be a solution that can deliver sufficient amount of data for training while the privacy of the data owners is respected.

To tackle this challenge, Google proposed federated learning~\cite{mcmahan2017communication} in 2016. Federated learning is a variation of distributed machine learning techniques that enables model training on a highly distributed client devices network. The key feature of federated learning is the training of models using the data collected locally, without transferring the data out of the client devices. A global model is initialised on a central server and broadcast to the participating client devices for local training. The locally trained model parameters are then collected by the central server and aggregated to update global model parameters. The global model parameters are broadcast again for the next training round. Each local training round usually takes a step in the gradient descent process. Fig.~\ref{FL_Overview} presents an overview of the federated learning process.

A federated learning system can be viewed as a large-scale distributed system, involving different components and stakeholders with diverse requirements and constraints. Hence, developing a federated learning system requires both software system design thinking and machine learning knowledge~\cite{lo2021architectural}. Further, despite having various reference architectures for machine learning, big data, industrial IoT, and edge computing systems, to the best of our knowledge, there is still no reference architecture for an end-to-end federated learning system. Based on findings in several federated learning reviews~\cite{10.1145/3450288,kairouz2019advances}, the application of federated learning is still limited and immature, with only certain stages of an end-to-end federated learning architecture are extensively studied, leaving many unfilled gaps for architecture and pipeline development. In contrast, many reusable solutions and components were proposed to solve the different challenges of federated learning systems and this motivates the design of a general federated learning system reference architecture. Therefore, this paper presents a pattern-oriented reference architecture that serves as an architecture design guideline and to facilitate the end-to-end development and operations of federated learning systems, while taking different quality attributes and constraints into considerations. This work provides the following contributions:

\begin{itemize}

    \item A pattern-oriented federated learning reference architecture named FLRA, generated from the findings of a systematic literature review (SLR) and mining of industrial best practices on machine learning system implementations.
    \item A pool of patterns associated with the different components of the FLRA reference architecture that target to address the recurring design problems in federated learning architectures.
    
\end{itemize}

The structure of the paper is as follows. Section~\ref{methodology} introduces the methodology for the reference architecture design, followed by the presentation of the reference architecture in Section~\ref{RA}. Section~\ref{Related works} presents the related work. Section~\ref{Conclusion} presents the discussions of this work and finally concludes this paper. 

\section{Methodology} \label{methodology}

We have employed parts of an empirically-grounded design methodology~\cite{10.1145/2000259.2000285} to design the federated learning reference architecture. Firstly, the design and development of this reference architecture are based on empirical evidence collected through our systematic literature review on 231 federated learning academic literature from a software engineering perspective~\cite{10.1145/3450288}. The review is conducted based on Kitchenham’s guideline~\cite{kitchenham2009systematic} with which we designed a comprehensive protocol for the review's initial paper search, paper screening, quality assessments, data extractions, analyses, and synthesis. We have also adopted the software development practices of machine learning systems in~\cite{8812912} to describe the software development lifecycle (SDLC) for federated learning. Using the stages of this lifeycycle as a guide, we formulated our research questions as: (1) Background understanding; (2) Requirement analysis; (3) Architecture design; and (4) Implementation \& evaluation. One major finding of the SLR is that federated learning research and applications are still highly immature, and certain stages of an end-to-end federated learning architecture still lack extensive studies~\cite{10.1145/3450288}. However, we have also identified many solutions and components proposed to solve the different challenges of federated learning systems, which can be reused and adapted.  This motivates the design of a federated learning system reference architecture.

\begin{figure}[t]
\includegraphics[width=\textwidth]{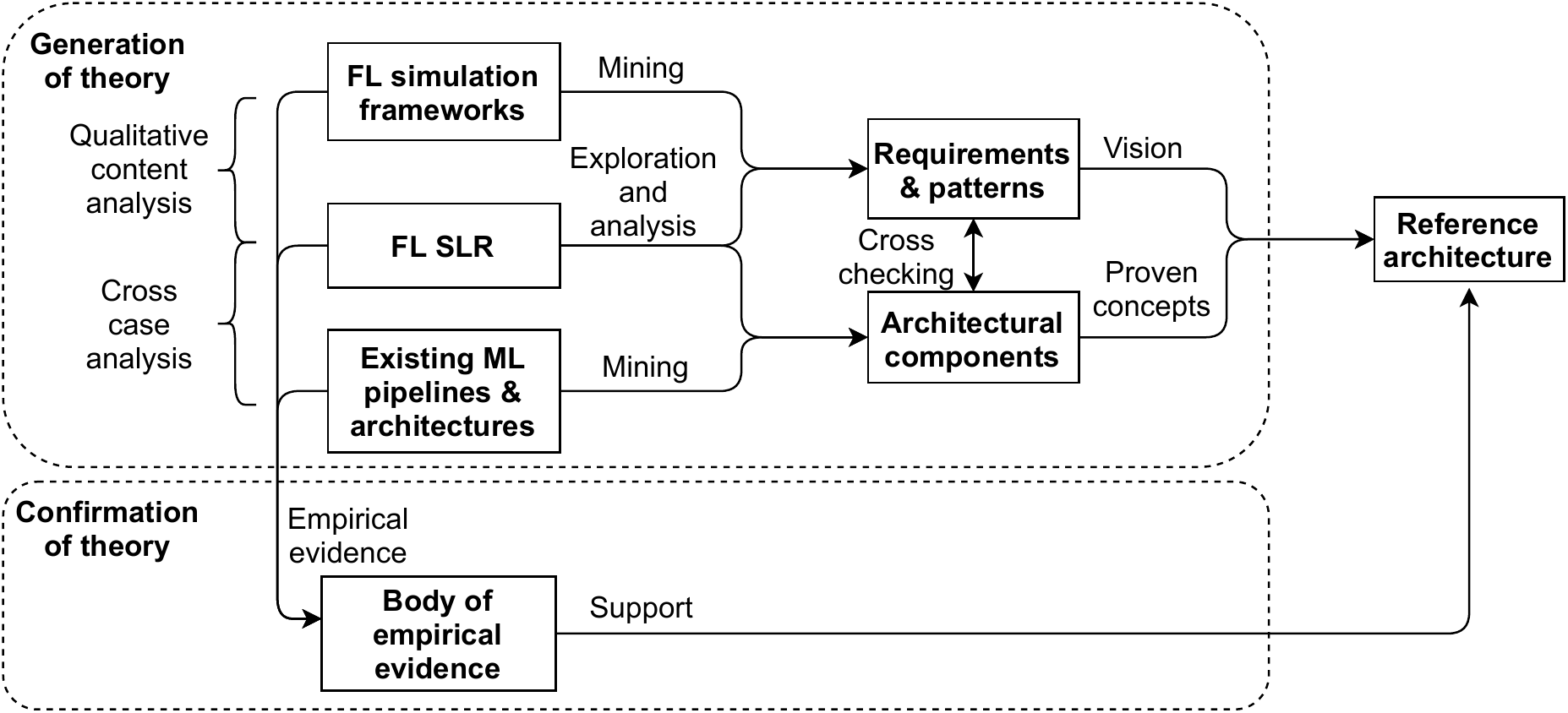}
\caption{Methodology for federated learning reference architecture design.} \label{FL_theory}
\end{figure}

Based on the findings, we specifically adopted the qualitative methods in empirical studies of software architecture~\cite{799955} to develop and confirm the theory for the reference architecture design. The proposition is generated based on syntheses and validations of the different recurring customers and business needs of the federated learning systems, in addition to the collections and analyses of the reusable patterns to address these architectural needs. We then conducted studies on some of the best practices in centralised and distributed machine learning systems to cover some of the components that are not covered in the federated learning studies. The main processes are the: (1) \textit{generation of theory} and (2) \textit{confirmation of theory}. The architecture design methodology is illustrated in Fig.~\ref{FL_theory}.

\subsection{Generation of theory}
The generation of the initial design of the reference architecture theory is performed in this stage. Since there is no standard reference architecture for federated learning yet, we generated the theory by referring to the architecture of a machine learning system. Here, we adopted \textit{cross-case analysis}~\cite{799955} as the theory generation method, which is an analysis method that compares two different cases based on some attributes and examines their similarities and differences. We performed a \textit{cross-case analysis} on the pipeline design of conventional machine learning and federated learning systems. Here, we reviewed several machine learning architectures proposed by well-known companies, such as Google\footnote{\url{https://cloud.google.com/architecture/mlops-continuous-delivery-and-automation-pipelines-in-machine-learning}}, Microsoft\footnote{\url{https://docs.microsoft.com/en-us/azure/machine-learning/concept-model-management-and-deployment}}, and Amazon\footnote{\url{https://docs.aws.amazon.com/sagemaker/latest/dg/multi-model-endpoints.html}}, specifically on their machine learning pipeline designs. Furthermore, based on our previous project implementation experience, we defined a general federated learning pipeline based on the standards proposed by these industry players that covers \textit{job creation, data collection, data preprocessing (cleaning, labeling, augmentation, etc.), model training, model evaluation, model deployment}, and \textit{model monitoring} stage. Since federated learning was first introduced by Google, the pipeline components analysis and mining are performed heavily on the federated learning standards proposed by Google researchers in~\cite{mcmahan2017communication,bonawitz2019towards,kairouz2019advances}, and the frameworks for federated learning system benchmark and simulation, such as Tensorflow Federated (TFF)\footnote{\url{https://www.tensorflow.org/federated}}, LEAF\footnote{\url{https://github.com/TalwalkarLab/leaf}}, and FedML\footnote{\url{https://fedml.ai/}}. From the findings, we were able to conclude that the \textit{data collection} is fairly similar whereas \textit{data preprocessing, model training, model evaluation, model deployment}, and \textit{model monitoring} stages for federated learning systems are different from machine learning pipelines. Especially for the \textit{model training} stage, the federated learning pipelines encapsulate \textit{model broadcast, local model training, model upload and collection}, and \textit{model aggregation} operation under this single stage. Furthermore, the iterative interaction between multiple client devices with one central server is the key design consideration of the federated learning architecture, and therefore, most academic work extensively studied the \textit{model training} stage and proposed many solutions which can be adapted as reusable components or patterns to address different requirements. 

Besides observing the pipeline design and the components, we performed \textit{qualitative content analyses} on existing machine learning and federated learning systems proposed by industrial practitioners and academics to extract requirements, reusable patterns, and components for the design of the reference architecture. In the SLR, a series of system quality attributes are defined based on ISO/IEC 25010 System and Software Quality model\footnote{\url{https://iso25000.com/index.php/en/iso-25000-standards/iso-25010}} and ISO/IEC 25012\footnote{\url{https://iso25000.com/index.php/en/iso-25000-standards/iso-25012}} Data Quality model to record the different challenges of a federated learning system addressed by researchers. The empirical evidence associated with each quality attribute and business need is analysed and validated as the support for the design proposition of the reference architecture. After the generation of theory for the hypothesis of the reference architecture, we designed the reference architecture according to the theory.

\subsection{Confirmation of theory}
In this stage, we confirmed and verified the viability and applicability of the reference architecture proposed. Since this reference architecture is built from scratch based on the patterns and requirements collected through qualitative analyses, we evaluated the architecture by building a convincing body of evidence to support the reference architecture, which is different from conventional evaluation approaches. We employed the qualitative validation method known as \textit{triangulation}~\cite{799955}. The basic idea is to gather different types of evidence to support a proposition. The evidence might come from different sources, be collected using different methods, and in our case, the evidence is from the SLR and the industrial implementations of machine learning systems from renowned institutions and companies, and our previous implementation experience. 

We have reviewed these evidence based on the SLDC lifecycle of machine learning systems we developed for the SLR to identify the adoptions of the different reusable patterns or components, in addition to the basic machine learning pipeline components that are mentioned in these evidence. These mentions and adoptions are collected to prove applicability of the instantiated components in the federated learning reference architecture. In short, the \textit{triangulation} process justified that the reference architecture is applicable as it is supported by various empirical evidence we collected and analysed.

\begin{figure}
\includegraphics[width=\textwidth]{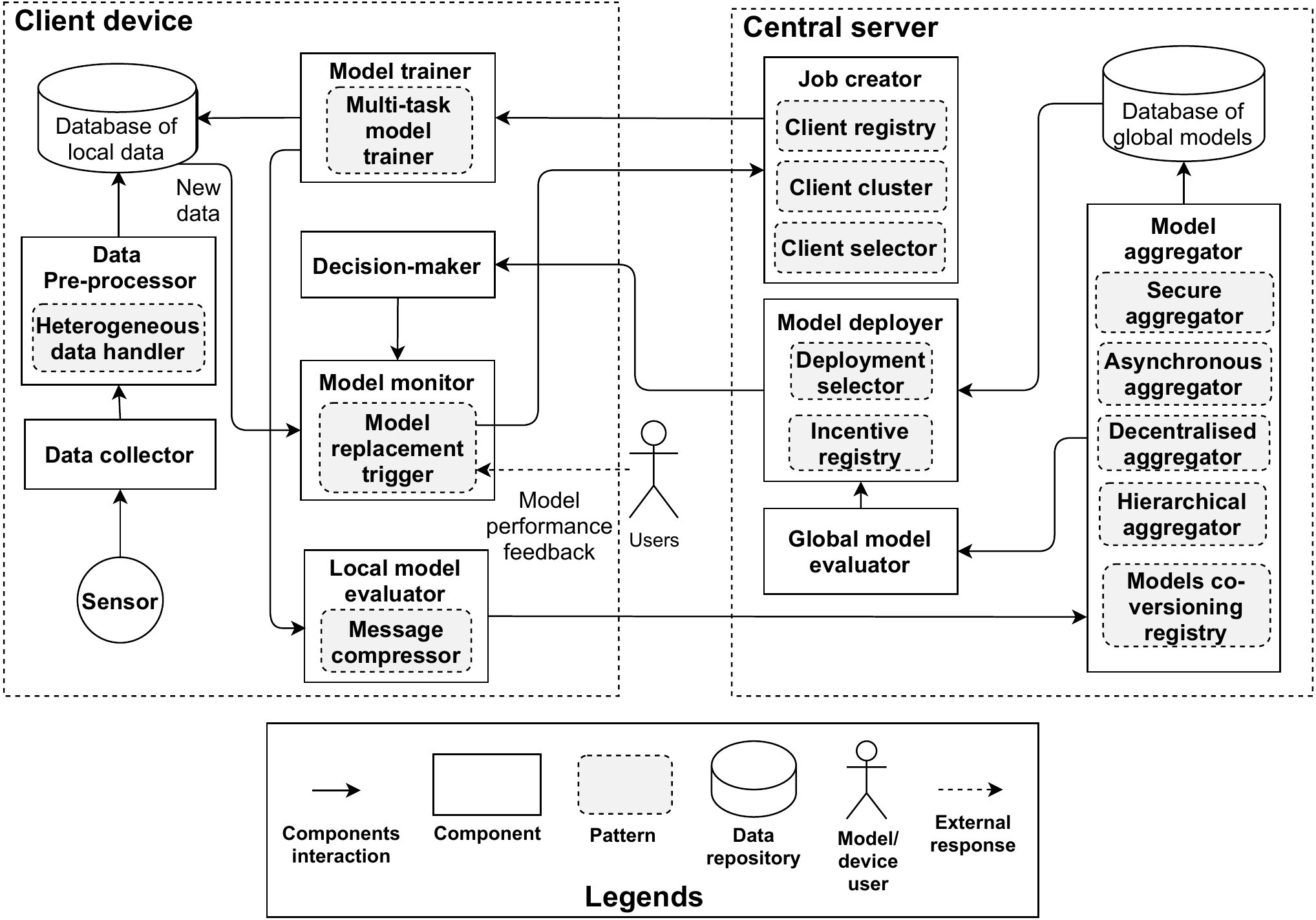}
\caption{FLRA: a reference architecture of federated learning systems.} \label{FL_RA}
\end{figure}

\section{FLRA Reference Architecture} \label{RA}
In this section, we present FLRA, a pattern-oriented reference architecture for federated learning systems. Fig.~\ref{FL_RA} illustrates the overall reference architecture. A base version of a federated learning system consists of two main participants: (1) central server and (2) client devices. A central server initialises a machine learning job and coordinates the federated training process, whereas client devices perform model training using local data and computation resources. 

Underneath the two participants, there are two types of components: (1) Mandatory components and (2) optional components. The mandatory components provide the basic functions required by a federated machine learning pipeline.
To fulfill the different software quality requirements and design constraints in federated learning systems, we collected and defined a set of patterns based on our SLR results and the mining of some existing federated learning simulation frameworks. Each pattern is embedded as optional components to facilitate the architecture design.

We summarised all the mandatory and optional components of the reference architecture and briefly highlighted the functionalities and responsibility of each component in Table~\ref{tab:Component_overview}. The table presents the details of each component associated with the federated learning pipeline stages.

\begin{table*}[tbp]
\scriptsize
\centering
\caption{Components of the federated learning reference architecture}
\label{tab:Component_overview}
\begin{tabular}{p{0.2\columnwidth}p{0.15\columnwidth}p{0.2\columnwidth}p{0.45\columnwidth}}
\toprule

\scriptsize\textbf{\makecell[l]{Stages}} & 
\scriptsize\textbf{\makecell[l]{Types}} &
\scriptsize\textbf{\makecell[l]{Components}} &
\scriptsize\textbf{\makecell[l]{Responsibility}}\\
\midrule

\multirow{4}{0.2\columnwidth}{{\\Job creation}} & \multirow{1}{0.15\columnwidth}{{Mandatory}} & \multirow{1}{0.2\columnwidth}{{Job creator}} & Initialises training job and global model\\
\cmidrule(l){2-4}

& \multirow{3}{0.15\columnwidth}{Optional} & \multirow{1}{0.2\columnwidth}{Client registry} & Improves system's \textbf{maintainability} and \textbf{reliability} by maintaining client's information\\
\cmidrule(l){3-4}

& &\multirow{1}{0.2\columnwidth}{Client cluster} & Tackles \textbf{statistical heterogeneity} \& \textbf{system heterogeneity} by grouping clients with similar data distribution or resources before aggregation\\
\cmidrule(l){3-4}

& &\multirow{1}{0.2\columnwidth}{Client selector} & Improves \textbf{model \& system's performance} by selecting high performance client devices\\

\cmidrule(l){1-4}
\cmidrule(l){1-4}

\multirow{3}{0.2\columnwidth}{\\Data\\collection\\\&\\preprocessing} & \multirow{2}{0.15\columnwidth}{Mandatory} & \multirow{1}{0.2\columnwidth}{Data collector} & Collects raw data through sensors or smart devices deployed\\
\cmidrule(l){3-4}

& &\multirow{1}{0.2\columnwidth}{Data preprocessor} & Preprocesses raw data \\
\cmidrule(l){2-4}

& \multirow{1}{0.15\columnwidth}{Optional} & \multirow{1}{0.2\columnwidth}{Heterogeneous Data Handler} & Tackles \textbf{statistical heterogeneity} through data augmentation methods\\

\cmidrule(l){1-4}
\cmidrule(l){1-4}

\multirow{10}{0.2\columnwidth}{\\Model\\training} & \multirow{3}{0.15\columnwidth}{Mandatory} & \multirow{1}{0.2\columnwidth}{Model trainer} & Trains local model\\
\cmidrule(l){3-4}

& &\multirow{1}{0.2\columnwidth}{Local model evaluator} & Evaluates local model performance after each local training round \\
\cmidrule(l){3-4}

& &\multirow{1}{0.2\columnwidth}{Model aggregator} & Aggregates local models to produce new global model \\
\cmidrule(l){2-4}

& \multirow{7}{0.15\columnwidth}{Optional} & \multirow{1}{0.2\columnwidth}{Multi-task\\model trainer} & Improves \textbf{model performance} (personalisation) by adopting multi-task training methods \\
\cmidrule(l){3-4}

& &\multirow{1}{0.2\columnwidth}{Message compressor} & Improves \textbf{communication efficiency} through message size reduction to reduce bandwidth consumption\\
\cmidrule(l){3-4}

& &\multirow{1}{0.2\columnwidth}{Secure aggregator} & Improves \textbf{data privacy} \& \textbf{system security} through different secure multiparty computation protocols \\
\cmidrule(l){3-4}

& &\multirow{1}{0.2\columnwidth}{Asynchronous aggregator} & Improves \textbf{system performance} by reducing aggregation pending time of late client updates\\
\cmidrule(l){3-4}

& &\multirow{1}{0.2\columnwidth}{Decentralised aggregator} & Improves system \textbf{reliability} through the removal of single-point-of-failure\\
\cmidrule(l){3-4}

& &\multirow{1}{0.2\columnwidth}{Hierarchical aggregator} & Improves \textbf{system performance} \& tackle \textbf{statistical heterogeneity} \& \textbf{system heterogeneity} by aggregating models from similar clients before global aggregation\\
\cmidrule(l){3-4}

& &\multirow{1}{0.2\columnwidth}{Model co-versioning\\registry} & Improves system's \textbf{accountability} by recording the local models associated to each global models to track clients' performances \\

\cmidrule(l){1-4}
\cmidrule(l){1-4}

\multirow{4}{0.2\columnwidth}{Model\\deployment} & \multirow{2}{0.15\columnwidth}{Mandatory} & \multirow{1}{0.2\columnwidth}{Model deployer} & Deploys completely-trained-models\\
\cmidrule(l){3-4}

& &\multirow{1}{0.2\columnwidth}{Decision maker} & Decides model deployment \\
\cmidrule(l){2-4}

& \multirow{4}{0.15\columnwidth}{Optional} & \multirow{1}{0.2\columnwidth}{Deployment selector} & Improves \textbf{model performance} (personalisation) through suitable model users selection according to data or applications\\
\cmidrule(l){3-4}

& &\multirow{1}{0.2\columnwidth}{Incentive registry} & Increases clients' \textbf{motivatability} \\

\cmidrule(l){1-4}
\cmidrule(l){1-4}

\multirow{2}{0.2\columnwidth}{\\Model\\monitoring} & \multirow{1}{0.15\columnwidth}{Mandatory} & \multirow{1}{0.2\columnwidth}{Model monitor} & Monitors model's data inference performance\\
\cmidrule(l){2-4}

&\multirow{1}{0.15\columnwidth}{Optional} & \multirow{1}{0.2\columnwidth}{Model replacement\\trigger} & Maintains \textbf{system \& model performance} by replacing outdated models due to performance degrades\\

\bottomrule
\end{tabular}
\end{table*}

\subsection{Job creation}
The federated learning process starts with the creation of a model training job (including initial model and training configurations) via \textit{\textbf{job creator}} on the central server. Within the \textit{\textbf{job creator}} component, three optional components could be considered are: \textit{\textbf{client registry}}, \textit{\textbf{client cluster}}, \textit{\textbf{client selector}}. In a federated learning system, client devices may be owned by different parties, constantly connect and drop out from the system. Hence, it is challenging to keep track of all the participating client devices including dropout devices and dishonest devices. This is different from distributed or centralised machine learning systems in which both clients and the server are typically owned and managed by a single party~\cite{MAROZZO20121382}. A \textit{\textbf{client registry}} is required to maintain all the information of the client devices that are registered, (e.g., ID, resource information, number of participating rounds, local model performance, etc.) Both IBM Federated Learning Framework\footnote{\url{https://github.com/IBM/federated-learning-lib}} and doc.ai\footnote{\url{https://doc.ai/}} adopted client registry in their design to improve maintainability and reliability of the system since the system can manage the devices effectively and quickly identify the problematic ones via the \textit{\textbf{client registry}} component. FedML which is a federated learning benchmarking and simulation framework has also explicitly covered the client manager module in their framework that serves the same purpose as the \textit{\textbf{client registry}}. However, the system may sacrifice client data privacy due to the recording of the device information on the central server.

The non-IID\footnote{Non-Identically and Independently Distribution: Highly-skewed and personalised data distribution that vary heavily between different clients and affects the model performance and generalisation~\cite{8889996}.} data characteristics of local raw data and the data-sharing restriction translates to model performance challenge~\cite{kairouz2019advances,mcmahan2017communication,zhao2018federated,li2019convergence}. When the data from client devices are non-IID, the global models aggregated is less generalised to the entire data. To improve the generalisation of the global model and speed up model convergence, a \textit{\textbf{client cluster}} component can be added to cluster the client devices into groups according to their data distribution, gradient loss, and feature similarities. This design has been used in Google's IFCA algorithm\footnote{\url{https://github.com/felisat/clustered-federated-learning}}, TiFL system\cite{10.1145/3369583.3392686}, and Massachusetts General Hospital's patient system\cite{HUANG2019103291}. The side effect of \textit{\textbf{client cluster}} is the extra computation cost caused by client relationship quantification.

The central servers interacts with a massive number of client devices that are both system heterogeneous and statistically heterogeneous. The magnitude of client devices number is also several times larger than that of the distributed machine learning systems~\cite{kairouz2019advances,10.1145/3450288}. To increase the model and system performance, client devices can be selected every round with predefined criteria (e.g., resource, data, or performance) via \textit{\textbf{client selector}} component. This has been integrated into Google's FedAvg~\cite{mcmahan2017communication} algorithm and IBM's Helios~\cite{xu2019helios}.

\subsection{Data collection \& preprocessing}
Each client device gathers data using different \textit{\textbf{sensors}} through the \textit{\textbf{data collector}} component and process the data (i.e., feature extraction, data cleaning, labeling, augmentation, etc.) locally through the \textit{\textbf{data preprocessor}} component, due to the data-sharing constraint. This is different from centralised or distributed machine learning systems in which the non-IID data characteristic is negligible since the data collected on client devices are usually shuffled and processed on the central server. Thus, within the \textit{\textbf{data preprocessor}}, an optional component \textit{\textbf{heterogeneous data handler}} is adopted to deal with the non-IID and skewed data distribution issue through data augmentation techniques. The known uses of the component include Astraea\footnote{\url{https://github.com/mtang724/Self-Balancing-Federated-Learning}}, FAug scheme~\cite{jeong2018communication} and Federated Distillation (FD) method~\cite{8904164}. 

\subsection{Model training}
\subsubsection{Local model training.} Once the client receives the job from the central server, the \textit{\textbf{model trainer}} component performs model training based on configured hyperparameters (number of epochs, learning rate, etc.). In the standard federated learning training process proposed by McMahan in~\cite{mcmahan2017communication}, only model parameters (i.e., weight/gradient) are mentioned to be sent from the central server, whereas in this reference architecture, the models include not only the model parameters but also the hyperparameters. For multi-task machine learning scenarios, a \textit{\textbf{multi-task model trainer}} component can be chosen to train task-related models to improve model performance and learning efficiency. Multitask Learning is a machine learning approach to transfer and share knowledge through training of individual models. It improves model generalisation by using the domain information contained in the parameters of related tasks as an inductive bias. It does this by learning tasks in parallel while using a shared representation; what is learned for each task can help other tasks be learned better~\cite{Caruana1998}. In federated learning scenarios, this technique is particularly relevant when faced with non-IID data which can produce personalised model that may outperform the best possible shared global model~\cite{kairouz2019advances}. This best practice solution is identified based on Google's MultiModel\footnote{\url{https://ai.googleblog.com/2017/06/multimodel-multi-task-machine-learning.html}} architecture, and Microsoft's MT-DNN\footnote{\url{https://github.com/microsoft/MT-DNN}}.

\subsubsection{Model evaluation.} The \textit{\textbf{local model evaluator}} component measures the performance of the local model and uploads the model to the \textit{\textbf{model aggregator}} on the central server if the performance requirement is met. In distributed machine learning systems, the performance evaluation on client devices is not conducted locally, and only the aggregated server model is evaluated. However, for federated learning systems, local model performance evaluation is required for system operations such as client selection, model co-versioning, contributions calculation, incentive provision, client clustering, etc.

\subsubsection{Model uploading.} The trained local model parameters or gradients are uploaded to the central server for model aggregation. Unlike centralised machine learning systems that performs model training in a central server or distributed machine learning systems that deals with fairly small amount of client nodes, the cost for transmitting model parameters or gradients between the bandwidth-limited client devices and central server is high when the system scales up~\cite{kairouz2019advances,10.1145/3450288}. A \textit{\textbf{message compressor}} component can be added to improve communication efficiency. The embedded pattern are extracted from Google Sketched Update~\cite{konevcny2016federated}, and IBM PruneFL~\cite{jiang2019model}.

\subsubsection{Model aggregation.} The \textit{\textbf{model aggregator}} formulates the new global model based on the received local models. There are four types of aggregator-related optional components within the \textit{\textbf{model aggregator}} component: \textit{\textbf{secure aggregator}}, \textit{\textbf{asynchronous aggregator}}, \textit{\textbf{decentralised aggregator}}, and \textit{\textbf{hierarchical aggregator}}. A \textit{\textbf{secure aggregator}} component prevents adversarial parties from accessing the models during model exchanges through multiparty computation protocols, such as differential privacy or cryptographic techniques. These techniques provide security proof to guarantee that each party knows only its input and output. For centralised and distributed machine learning settings that practice centralised system orchestration, communication security between clients and server is not the main concern. In contrast, for federated learning settings, this best practices are used in SecAgg~\cite{10.1145/3133956.3133982}, HybridAlpha~~\cite{10.1145/3338501.3357371}, and TensorFlow Privacy Library\footnote{\url{https://github.com/tensorflow/privacy/}}. \textit{\textbf{Asynchronous aggregator}} is identified from ASO-fed~\cite{chen2020asynchronous}, AFSGD-VP~\cite{gu2020privacy}, and FedAsync~\cite{xie2019asynchronous}. The \textit{\textbf{asynchronous aggregator}} component enables the global model aggregation to be conducted asynchronously whenever a local model update arrives. Similar technique have been adopted in distributed machine learning approaches such as iHadoop~\cite{6133130} and it is proven that this can effectively reduce the overall training time. The conventional design of a federated learning system that relies on a central server to orchestrate the learning process might lead to a single point of failure. A \textit{\textbf{decentralise aggregator}} performs model exchanges and aggregation in decentralised manner to improve system reliability. The known uses of \textit{\textbf{decentralised aggregator}} include BrainTorrent~\cite{roy2019braintorrent} and FedPGA~\cite{9205506}. Blockchain can be employed as a decentralised solution for federated learning systems. In distributed machine learning systems, p2p network topology is employed to in MapReduce~\cite{MAROZZO20121382} to resolve the single-point-of-failure threat on parameter servers. A \textit{\textbf{hierarchical aggregator}} component can be selected to improve system efficiency by adding an intermediate edge layer to aggregate the model updates from related client devices partially before performing the final global aggregation. This pattern has been adopted by HierFAVG~\cite{9148862}, Astraea, and HFL~\cite{9054634}.

In addition to aggregator-related optional components, a \textit{\textbf{model co-versioning registry}} component can be embedded within the \textit{\textbf{model aggregator}} component to map all the local models and their corresponding global models. This enables the model provernance and improves system accountability. The \textit{\textbf{model co-versioning registry}} pattern is summarised and adopted from the version control methods in DVC\footnote{\url{https://dvc.org}}, Replicate.ai\footnote{\url{https://replicate.ai}}, and Pachyderm\footnote{\url{https://www.pachyderm.com}}.

\subsection{Model deployment}
After the aggregation, the \textit{\textbf{global model evaluator}} assesses the performance of the global model. One example is TensorFlow Extended (TFX)\footnote{\url{https://www.tensorflow.org/tfx}} that provides a model validator function to assess the federated learning model performance. If the global model performs well, the \textit{\textbf{model deployer}} component deploys the global model to the client device for decision-making through the \textit{\textbf{decision-maker}} component. For instance, TensorFlow lite\footnote{\url{https://www.tensorflow.org/lite}} prepares the final validated model for deployment to the client devices for data inference. Within the \textit{\textbf{model deployer}} component, there are two optional components for selection: \textit{\textbf{deployment selector}} and \textit{\textbf{incentive registry}}. The \textit{\textbf{deployment selector}} component examines the client devices and selects clients to receive the global model based on their data characteristics or applications. The \textit{\textbf{deployment selector}} pattern has been applied in Azure Machine Learning\footnote{\url{https://docs.microsoft.com/en-us/azure/machine-learning/concept-model-management-and-deployment}}, Amazon SageMaker\footnote{\url{https://docs.aws.amazon.com/sagemaker/latest/dg/multi-model-endpoints.html}}, and Google Cloud\footnote{\url{https://cloud.google.com/ai-platform/prediction/docs/deploying-models}} to improve model performance. The {incentive registry} component maintains all the client devices' incentives based on their contributions and agreed rates to motivate clients to contribute to the training. Blockchain has been leveraged in FLChain~\cite{8905038} and DeepChain~\cite{8894364} to build a \textit{\textbf{incentive registry}}.

\subsection{Model monitoring}
After the deployment of models for the actual data inference, a \textit{\textbf{model monitor}} keeps track of the model performance continuously. If the performance degrades below a predefined threshold value, the \textit{\textbf{model replacement trigger}} component notifies the \textit{\textbf{model trainer}} for local fine-tuning or sends an alert to the \textit{\textbf{job creator}} for a new model generation. The \textit{\textbf{model replacement trigger}} pattern is identified based on the known uses including Microsoft Azure Machine Learning Designer\footnote{\url{https://azure.microsoft.com/en-au/services/machine-learning/designer}}, Amazon SageMaker\footnote{\url{https://aws.amazon.com/sagemaker}}, Alibaba Machine Learning Platform\footnote{\url{https://www.alibabacloud.com/product/machine-learning}}.

\section{Related Work} \label{Related works}
The most widely mentioned definition of a reference architecture is defined by Bass et al.~\cite{bass2003software} as ``a reference model mapped onto software elements (that cooperatively implement the functionality defined in the reference model) and the data flow between them. Whereas a reference model divides the functionality, a reference architecture is the mapping of that functionality onto a system decomposition.'' Nakagawa et al. collected a series of definitions of reference architectures by various researchers and summarised them as follows: ``the reference architecture encompasses the knowledge about how to design system architectures of a given application domain. It must address the business rules, architectural styles (sometimes also defined as architectural patterns that address quality attributes in the reference architecture), best practices of software development (architectural decisions, domain constraints, legislation, and standards), and the software elements that support the development of systems for that domain~\cite{10.1007/978-3-642-23798-0_22}.'' 

Reference architectures for machine learning applications and big data analysis were researched comprehensively. For instance, Pääkkönen and Pakkala proposed a reference architecture of big data systems for machine learning in an edge computing environment~\cite{paakkonen2020extending}. IBM AI Infrastructure Reference Architecture is proposed to be used as a reference by data scientists and IT professionals who are defining, deploying, and integrating AI solutions into an organization~\cite{lui_karmiol_2018}. 

Reference architectures for edge computing systems are also widely studied. For example, H2020 FAR-Edge-project, Edge Computing Reference Architecture 2.0, Intel-SAP Reference Architecture, IBM Edge computing reference architecture, and Industrial Internet Reference Architecture (IIRA) are proposed by practitioners to support the development of multi-tenant edge systems. 

There are existing works proposed to support federated learning system and architecture design. For instance, Google was the earliest to introduce a system design approach for federated learning~\cite{bonawitz2019towards}. A scalable production system for federated learning in the domain of mobile devices, based on TensorFlow described from a high-level perspective. A collection of architectural patterns for the design of federated learning systems are summarised and presented by~\cite{lo2021architectural}. There are also many architectures and adoptions of federated learning systems proposed by researchers for diverse applications. For instance, Zhang et al.~\cite{9233457} proposed a blockchain-based federated learning architecture for industrial IoT to improve client motivatability through an incentive mechanism. Samarakoon et al.~\cite{8647927} have adopted federated learning to improve reliability and communication latency for vehicle-to-vehicle networks. Another real-world federated learning adoption by Zhang et al.~\cite{9347454} is a dynamic fusion-based federated learning approach for medical diagnostic image analysis to detect COVID-19 infections. We observed that there have been multiple studies on federated learning from different aspects and their design methods are highly diverse and isolated which makes their proposals challenging to be reproduced. 

Motivated by the previous works mentioned above, we intend to fill the research gap by putting forward an end-to-end reference architecture for federated learning systems development and deployment which has been distinctly lacking in the current state-of-the-art.

\section{Discussion \& Conclusion} \label{Conclusion}
A reference architecture can be served as a standard guideline for system designers and developers for quick selection of best practice solutions for their problems, which can be further customised as required. To the best of our knowledge, there is still no reference architecture proposed for an end-to-end federated learning system while many reusable components and patterns have been proposed. Thus, in this paper, we proposed FLRA, a pattern-oriented reference architecture for federated learning system design to increase the real-world adoption of federated learning.

To design the reference architecture, we developed an empirically-grounded qualitative analysis method as the basis of design theory generation. The empirical evidence to support the reference architecture design is a collection of findings (requirements, patterns, and components) gathered and defined by our previous systematic literature review on federated learning and well-known industry practices of machine learning systems. 

After developing the reference architecture, we compared it with existing machine learning architectures of Google, Amazon, Microsoft, and IBM to examine its applicability. The key differences between centralised or distributed machine learning with federated learning are the non-IIDness of training data, variation in the data partitioning (e.g., vertical, horizontal, and transfer federated learning) and device partitioning (e.g., cross-device, cross-silo), the ownership and security requirements of different client devices, the system heterogeneity, and the participation of client nodes. The proposed FLRA architecture adopted many reusable machine learning and federated learning patterns while maintaining most of the mandatory machine learning pipeline components. This ensures that the reference architecture is generalised to support the basic model training tasks in the real world. 

While there are different constraints when developing a federated learning system for different applications and settings, the possible trade-offs and the pattern solutions to these challenges are discussed comprehensively. The confirmation of theory justified the applicability of FLRA and the patterns associated with the support of empirical evidence collected. Hence, the FLRA proposed is applicable in the real world for a general, end-to-end development of federated learning systems. Our future work will focus on developing an architecture decision model for federated learning system design. We will also work on the architecture design for trust in federated learning systems.

%
%
%
\bibliographystyle{splncs04}
\bibliography{mybibliography}

\end{document}